\documentclass[acmsmall]{acmart}
\usepackage[capitalise]{cleveref}
\usepackage{subcaption}
\usepackage{algorithm}
\usepackage{algpseudocode}
\usepackage{multirow}
\usepackage{pdfpages}
\usepackage{lipsum}
\usepackage{enumitem}
\setlist[enumerate]{itemsep=0mm}
\usepackage{soul}
\usepackage{xcolor}

\AtBeginDocument{%
  \providecommand\BibTeX{{%
    \normalfont B\kern-0.5em{\scshape i\kern-0.25em b}\kern-0.8em\TeX}}}

\setcopyright{acmlicensed}
\copyrightyear{2025}
\acmYear{2025}
\acmDOI{XXXXXXX.XXXXXXX}

\usepackage{fancyhdr}   

\fancypagestyle{firstpage}{     
  \fancyhf{}
  \chead{\textcolor{gray}{This article has been accepted by MobileHCI2025 Workshops}}
}

\begin{document}

\definecolor{Orange}{rgb}{0.8,0.4,0}
\definecolor{Green}{rgb}{0,0.5,0}
\definecolor{Blue}{rgb}{0,0,1}
\definecolor{Red}{rgb}{0.7,0,0}
\definecolor{Aqua}{rgb}{0,0.5,0.5}
\definecolor{airforceblue}{rgb}{0.36, 0.54, 0.66}
\definecolor{darkblue}{rgb}{0.0, 0.0, 0.55}
\definecolor{light-gray}{gray}{0.95}
\definecolor{burgundy}{RGB}{144,0,32}

\newcommand{\todo}[1]{}
\newcommand{\todofu}[1]{}
\newcommand{\todofig}[1]{\textsf{\textbf{\textcolor{Red}{[TODO FIG: #1]}}}}
\newcommand{\done}[1]{\textbf{\textcolor{Green}{[DONE: #1]}}}


\newcommand{\change}[1]{#1}
\newcommand{\changedelete}[1]{}


\newcommand{\dfkiauthor}[3]{
\author{#1}
\orcid {#3}
\email{#2}
\affiliation{%
  \institution{DFKI}
  \city{Kaiserslautern}
  \country{Germany}
}
}

\newcommand{\dfkianduniauthor}[3]{
\author{#1}
\orcid {#3}
\email{#2}
\affiliation{%
  \institution{DFKI and RPTU}
  \city{Kaiserslautern}
  \country{Germany}
}
}


\newcommand{\ressec}[1]{\subsubsection*{$\square \square$ #1:}} 
\newcounter{takeawaycounter} \setcounter{takeawaycounter}{0}
\newcommand{\takeaway}{\subsubsection*{$\square \square$ Takeaway message \arabic{takeawaycounter}: } \stepcounter{takeawaycounter}}

\newcommand{\guideline}[1]{\subsubsection{#1}}

\newcommand{\insertfig}[5]{

\begin{figure}[t!] 
	\centering
    \includegraphics[width=#4\columnwidth]{#1}
    \caption{#3}   
    \label{#2}
    \Description{#5}
\end{figure}

}

\newcommand{\insertsubfig}[5]{
\centering
  \begin{subfigure}{#4\textwidth}
  \includegraphics[width=1.0\linewidth]{#1} 
  \caption{#3}
  \label{#2}
  \Description{#5}
  \end{subfigure}
}

\newcommand{\insertappendix}[3]{
\includepdf[scale=0.65,pages=1,pagecommand=\subsection{#1}\label{#3}]{#2}
}

\newcommand{\insertsurvey}[3]{
\includepdf[scale=0.7,pages=1,pagecommand=\section{#1}\label{#3}]{#2}
\includepdf[scale=0.8,pages=2-,pagecommand={}]{#2}
}

\newcommand{\insertdoc}[3]{
\includepdf[scale=0.8,pages=1,pagecommand=\section*{#1}\label{#3}]{#2}
\includepdf[scale=0.9,pages=2-,pagecommand={}]{#2}
}

\newcommand{\missingpax}[2]{We analyzed the data from #1 participants after excluding incomplete data from #2 participant(s).}
\newcommand{\tablestats}[3]{See Table \ref{#1} (row #2 to #3) for details about the statistics.}
\newcommand{\cone}{C1~}
\newcommand{\ctwo}{C2~}
\newcommand{\cthree}{C3~}

\newcommand{\meansd}[2]{$(Mean= #1, SD= #2)$}
\newcommand{\mean}[1]{$(M= #1)$}
\newcommand{\mrng}[1]{(means $\leq$ #1)}
\newcommand{\mrngbig}[1]{(means $\geq$ #1)}
\newcommand{\bfval}[1]{$(BF_{10}=#1)$}
\newcommand{\bfnull}{$BF_{10}$}


\title{Bridging Generalization and Personalization in Human Activity Recognition via On-Device Few-Shot Learning}

\author{Pixi Kang}
\email{kang.pixi@sanechips.com.cn}
\orcid{0000-0003-1191-5815}
\affiliation{%
  \institution{Tsinghua University, Sanechips Technology Co.,Ltd.}
  \country{China}
}

\author{Julian Moosmann}
\email{julian.moosmann@pbl.ee.ethz.ch}
\orcid{0009-0007-0283-0031}
\affiliation{%
  \institution{ETH Zürich}
  \country{Switzerland}
}

\dfkiauthor{Mengxi Liu}{mengxi.liu@dfki.de}{0000-0003-0527-1208}
\dfkianduniauthor{Bo Zhou}{bo.zhou@dfki.de}{0000-0002-8976-5960}

\author{Michele Magno}
\email{michel.magno@pbl.ee.ethz.ch}
\orcid{0000-0003-0368-8923}
\affiliation{%
  \institution{ETH Zürich}
  \country{Switzerland}
}

\dfkianduniauthor{Paul Lukowicz}{paul.lukowicz@dfki.de}{0000-0003-0320-6656}
\dfkianduniauthor{Sizhen Bian}{sizhen.bian@dfki.de}{0000-0001-6760-5539}

\renewcommand{\shortauthors}{Kang et al.}

\begin{abstract}

Human Activity Recognition (HAR) with different sensing modalities requires both strong generalization across diverse users and efficient personalization for individuals. However, conventional HAR models often fail to generalize when faced with user-specific variations, leading to degraded performance. To address this challenge, we propose a novel on-device few-shot learning framework that bridges generalization and personalization in HAR. Our method first trains a generalizable representation across users and then rapidly adapts to new users with only a few labeled samples, updating lightweight classifier layers directly on resource-constrained devices. This approach achieves robust on-device learning with minimal computation and memory cost, making it practical for real-world deployment. We implement our framework on the energy-efficient RISC-V GAP9 microcontroller and evaluate it on three benchmark datasets (RecGym, QVAR-Gesture, Ultrasound-Gesture). Across these scenarios, post-deployment adaptation improves accuracy by 3.73\%, 17.38\%, and 3.70\%, respectively. These results demonstrate that few-shot on-device learning enables scalable, user-aware, and energy-efficient wearable human activity recognition by seamlessly uniting generalization and personalization. The related framework is open sourced for further research
\footnote{https://github.com/kangpx/onlineTiny2023}.

\end{abstract}

\ccsdesc[500]{Human-centered computing~Interaction design}
\ccsdesc[500]{Human-centered computing}

\keywords{Generalization, Personalization, Human Activity Recognition, On-Device Learning, Few-Shot Learning}

\received{06 February 2025}
\received[revised]{XX XX 2025}
\received[accepted]{XX XX 2025}

\maketitle

\section{Introduction}

\thispagestyle{firstpage} 

Wearable Human Activity Recognition (HAR) plays a crucial role in fitness tracking \cite{bian2019passive}, healthcare monitoring \cite{iqbal2021advances}, and smart assistance systems~\cite{saleem2023toward,singh2023recent}. These systems rely on a wide range of sensors embedded in consumer devices such as smartwatches, fitness bands, and body-worn nodes to collect time-series data related to movement \cite{bian2025magnetic}, posture \cite{salaorni2024wearable}, and gestures \cite{liu2025ibreath}. As deep learning methods have grown in popularity, many HAR models have adopted powerful architectures trained on large-scale datasets to generalize across multiple users and contexts~\cite{hou2020study,ordonez2016deep}.
However, a fundamental limitation remains: models trained on a broad population often underperform when deployed to new individuals \cite{mohamed2024real, hokka2024gender}. This discrepancy is primarily caused by user-induced concept drift (UICD)~\cite{mathur2020scaling,bian2022contribution}, where each user exhibits unique behavioral signatures, sensor placements, and body mechanics that deviate from the training distribution. This observation is consistent with recent findings on the scaling laws in HAR \cite{hoddes2025scaling}, which show that adding more users to a dataset yields greater performance gains than simply increasing the amount of data per user. As a common result, the accuracy of generalized HAR models degrades significantly in real-world deployments, leading to a poor user experience and limiting the scalability of such solutions \cite{hokka2024gender,holko2022wearable}.

To address this issue, we advocate a two-stage approach: generalize first using full data to build a broadly applicable model, and then personalize rapidly using few-shot learning directly on the device. This strategy offers a path to robust, user-aware inference while respecting hardware limitations and privacy constraints. We focus on enabling such adaptation on resource-constrained microcontrollers, specifically using the RISC-V GAP9 platform, and evaluate the solution with RecGym~\cite{bian2022exploring}, a dataset that combines Inertial Measurement Unit (IMU) and human body capacitance (HBC) signals to capture diverse gym activity patterns.
This work makes the following key contributions:

\begin{itemize}
    \item We quantify the impact of user-induced concept drift in wearable HAR across three representative scenarios. Using the RecGym, QVAR, and Ultra datasets, we observe accuracy drops of 0.74\%, 25.89\%, and 6.00\%, respectively, when generalizing to unseen users.
    \item We propose an efficient few-shot on-device learning strategy that updates only the dense classifier layer using streaming samples from the target user, allowing rapid personalization with minimal memory overhead.
    \item We implement a real-time on-device training engine optimized for the GAP9 microcontroller, enabling adaptation with a latency of sub-ms and energy consumption of $\mu$J-level per parameter update.
    \item We demonstrate that our hybrid approach improves average recognition accuracy post-deployment by 3.73\% (RecGym), 17.38\% (QVAR), and 3.70\% (Ultra), validating its robustness and effectiveness across diverse sensing modalities and activity types.
\end{itemize}

\section{Related Work}

Sensor-based human activity recognition (HAR) has primarily relied on inertial measurement units (IMUs) and camera-based methods~\cite{hou2020study,ordonez2016deep}, which offer rich information but face challenges such as accumulated drift, occlusion, and privacy concerns. Novel sensing modalities such as radar~\cite{zhou2024kolmogorovarnold}, ultrasound~\cite{ling2022ultragesture}, and electrostatic charge variation sensors~\cite{reinschmidt2022realtime} have been proposed to enhance robustness, but these systems remain sensitive to inter-user variability and often require heavy preprocessing or domain-specific model design.

To tackle user-induced concept drift (UICD), transfer learning and meta-learning approaches have been explored. Works such as~\cite{mathur2020scaling,an2023transfer} apply domain adaptation techniques to bridge the gap between training and deployment users, but these methods typically assume access to labeled data during retraining and lack practical on-device implementations. Other studies like~\cite{craighero2023ondevice,llisterri2022ondevice} investigate on-device adaptation using MCUs, but they either focus on simplified HAR scenarios or limit training to very constrained models, affecting final accuracy.

Few-shot adaptation has emerged as a promising paradigm to address personalization under low-resource settings. Solutions like TinyOL~\cite{ren2021tinyol} and DaCapo~\cite{khan2023dacapo} enable continuous learning on microcontrollers but are primarily validated on vision or speech tasks. In~\cite{ravaglia2021tinyml}, a continual learning framework using quantized latent replay is proposed, yet the method requires additional memory for buffer storage and is not evaluated on real-time wearable applications.

From a hardware perspective, platforms like STM32, MAX78000~\cite{analog2023max78000}, and GAP9~\cite{bian2022exploring} offer increasing support for AI workloads, including CNN accelerators and parallel compute clusters. However, few efforts combine hardware-efficient deployment with real-time adaptation tailored to sensor-based HAR.

In summary, while prior work has laid the foundation for model personalization and on-device learning, gaps remain in implementing efficient on-device adaptation pipelines for wearable HAR, and quantifying the trade-offs between generalization, personalization, and energy-latency efficiency in realistic settings.
Our work addresses these limitations by demonstrating an end-to-end system for few-shot on-device user adaptation.

\section{Methodology}
\subsection{Network Design}
We employ a lightweight yet effective 1D Convolutional Neural Network (1D-CNN) tailored for multi-channel time-series data typical in wearable HAR. The model is structured in two distinct parts: a fixed feature extractor (backbone) and a trainable dense classifier for personalization.

The backbone begins with a 1D convolutional layer, followed by batch normalization, a ReLU activation, and dropout. This is succeeded by three residual blocks, each containing convolutional layers with shortcut connections to retain temporal information while reducing gradient vanishing. The output is then flattened and passed into the classifier.

The classifier is a single dense layer followed by a softmax activation. During on-device adaptation, only this layer is updated, allowing fast personalization with low memory footprint. This structure ensures that general features are preserved while user-specific variations can be learned in a few-shot manner without full retraining.

\begin{figure}[htbp]
\centering 
\subfloat[]{\includegraphics[width=0.6\columnwidth]{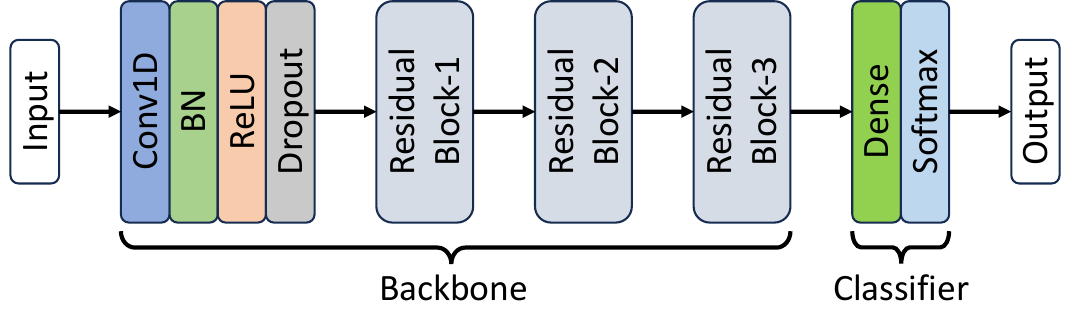}\label{fig:whole_network}}
\hfil
\subfloat[]{\includegraphics[width=0.6\columnwidth]{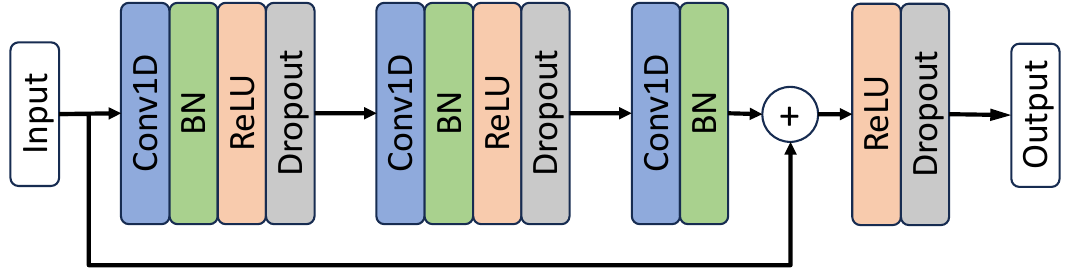}\label{fig:residual_block}}
\caption{Network topology. (a) Whole network. (b) Residual block.}
\label{fig:network_architecture}
\end{figure}

Figure~\ref{fig:network_architecture} visualizes this architecture. Such a separation of general and personal components is inspired by prior success in domain adaptation frameworks and is proven effective in our evaluation.


\subsection{On-Device Learning Engine}
\begin{figure*}[htbp]
\subfloat[]{
\begin{minipage}{0.675\linewidth}
    \centerline{\includegraphics[width=\textwidth]{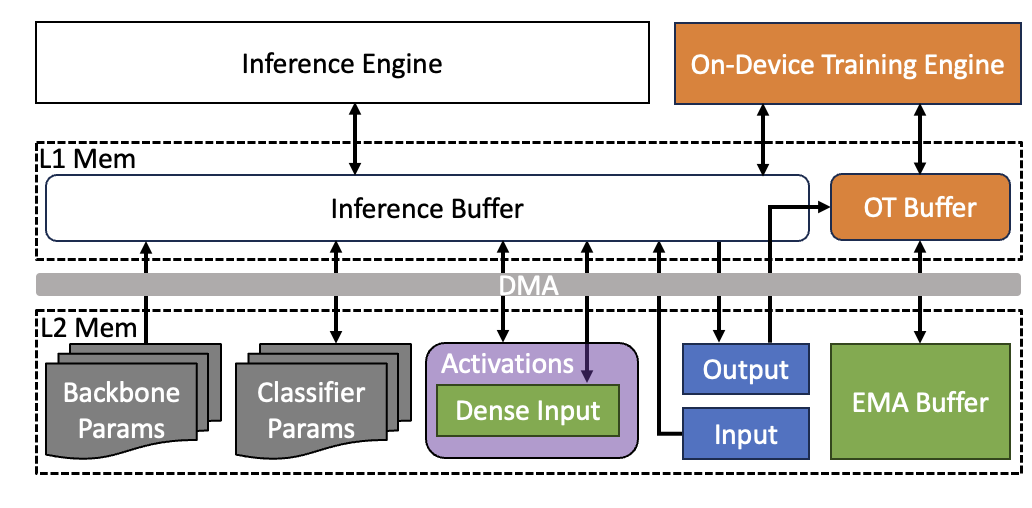}}
    \label{fig:ot-engine-stm32f7}
\end{minipage}
}

\caption{Architectures of implemented on-device training engines on GAP9 processor}
\label{fig:ot-engines}
\end{figure*}

To enable real-time user adaptation in constrained environments, we design and implement an on-device learning engine optimized for the GAP9 platform by GreenWaves Technologies. GAP9 is a low-power parallel processor that features a multi-core RISC-V cluster and dedicated hardware accelerators for signal processing and neural network workloads. Its memory hierarchy, which includes 1.6MB of L2 SRAM and 128KB of tightly coupled L1 memory, allows us to run training and inference tasks in a resource-efficient manner.
Our on-device engine is built around the idea of minimizing memory access and computational overhead while supporting efficient stochastic gradient descent (SGD) updates to the final classifier layer (Fig. \ref{fig:ot-engines}). At runtime, the backbone of the neural network remains frozen in L2 memory and is used solely for forward inference. The trainable dense layer, along with the associated exponential moving average (EMA) buffers used for momentum optimization, is stored in L1 memory to facilitate fast updates without frequent memory transfers.
When new labeled user samples are available, the GAP9 controller orchestrates the forward pass through the fixed backbone, computes gradients for the dense layer using cross-entropy loss, and performs parameter updates in-place. All memory movements between L2 and L1 are precompiled using AutoTiler scripts, and compute kernels are parallelized across the 8 RISC-V cores in the cluster. This multi-threaded update routine ensures that parameter adaptation is not only accurate but also latency-efficient.


\section{Experimental Setup}


\subsection{Datasets}
To evaluate the effectiveness and generality of our few-shot on-device personalization strategy, we use three publicly available sensor-based HAR datasets: RecGym, QVAR- and Ultra-Gesture. These datasets represent diverse sensing modalities and activity types, enabling a broad assessment of user-induced concept drift and adaptation effectiveness.
\textbf{RecGym\cite{bian2025hybrid}} captures multimodal data from wearable IMU and human body capacitance (HBC) during gym workouts. It contains recordings from 10 participants performing 12 structured exercises, collected via a wrist-worn device. 
\textbf{QVAR\cite{reinschmidt2022realtime}} is a hand gesture recognition dataset using QVAR sensor in combination with IMU. Data was collected from 20 users performing 10 hand gestures. 
\textbf{Ultra\cite{kangpx_40khz-ultrasonicdhgr-onlinesemi_2023}} contains ultrasonic-based hand gesture datasets includes 5,600 samples from 7 users performing 8 distinct dynamic gestures. 

\subsection{Preprocessing and Input Formatting}
The raw signals are segmented using a sliding window approach with a window size of 2 seconds and 50\% overlap, resulting in time-series tensors with shape $(C \times T)$ where $C$ is the number of channels and $T$ denotes time steps. Normalization is applied per channel to remove sensor-specific biases. These segments are used for both offline training and on-device personalization without any handcrafted feature extraction, ensuring the process remains fully learnable and end-to-end.

\subsection{Evaluation Protocol}
We employ a Leave-One-Person-Out (L1PO) cross-validation strategy to simulate real-world deployment. In each fold, data from one participant is excluded from training and reserved for evaluation. The model is first trained offline using the remaining users' data to simulate generalization. For adaptation, 40\% of the target user's samples are used for on-device training (few-shot), and the remaining 60\% are used for testing. This setup mirrors post-deployment usage where limited labeled samples become available for fast personalization.

\subsection{Quantifying User-Induced Concept Drift}
To motivate the need for on-device personalization, we quantify the effect of user-induced concept drift (UICD) across all three datasets: RecGym, QVAR, and Ultra. For each dataset, we compare model performance under two training conditions: a generalized setting using Leave-One-Person-Out (L1PO) and an oracle setting using Leave-One-Session-Out (L1SO), where data from the test user is included.

Our results show consistent performance degradation when generalizing to unseen users: RecGym exhibits a 0.74\% drop in accuracy, Ultra shows a 6.00\% drop, and QVAR experiences the most significant drop at 25.89\%. These findings highlight the variability in user-specific signal patterns across different sensing modalities and application contexts, reinforcing the necessity of lightweight personalization to restore performance in real-world deployments.

\section{Results and Discussion}

To comprehensively evaluate the effectiveness of our proposed few-shot personalization framework, we conduct experiments across three distinct HAR datasets—RecGym, QVAR, and Ultra—deployed on the GAP9 embedded platform.  We report both classification accuracy and system-level metrics before and after on-device training to assess model robustness, adaptation quality, and runtime efficiency.

\subsection{Accuracy Improvements Through Personalization}
In all three datasets, the general model was trained solely on non-user data using Leave-One-Person-Out (L1PO) cross-validation. For each target user, 40\% of their data was used for on-device few-shot personalization, while the remaining 60\% was reserved for testing. Table~\ref{tab:multi_dataset_accuracy} summarizes the average accuracy before and after adaptation for each dataset.

\begin{table}[h]
\centering
\caption{Average Recognition Accuracy Before and After On-Device Personalization}
\label{tab:multi_dataset_accuracy}
\begin{tabular}{|l|c|c|c|}
\hline
\textbf{Dataset} & \textbf{Pre-ODP Accuracy} & \textbf{Post-ODP Accuracy} & \textbf{Improvement} \\
\hline
RecGym & 90.37\% & 94.04\% & +3.73\% \\
QVAR & 67.13\% & 84.00\% & +17.38\% \\
Ultra & 91.93\% & 95.69\% & +3.70\% \\
\hline
\end{tabular}
\end{table}

The QVAR dataset exhibits the largest gain (+17.38\%), likely due to its high user-specific signal characteristics and greater inter-user variability. Ultra and RecGym show more moderate but consistent improvements, despite having higher baseline generalization performance. These results confirm that lightweight classifier adaptation can significantly recover or exceed baseline performance, even when the model was initially trained on unrelated users.

Across all datasets, personalization was achieved with a single forward-backward pass per sample. The backbone remained frozen, ensuring minimal compute and memory load—demonstrating the feasibility of this approach in real-world edge scenarios.

\subsection{Latency and Energy Efficiency}
To evaluate system responsiveness, we measured the average inference latency and parameter update latency using onboard timers. The inference for a single sample on GAP9 takes approximately 0.34 milliseconds, whereas each on-device training update (including forward pass, loss computation, and parameter adjustment) completes within 0.07–0.17 milliseconds, depending on input size.

In terms of energy, each inference consumes about 35 microjoules, while a full parameter update requires roughly 4 microjoules per sample. This translates to over 250× lower energy consumption for training updates compared to typical MCU-class devices, such as STM32F7, which consumes approximately 400 microjoules per update. The parallel architecture and memory-optimized training kernel of GAP9 are instrumental in achieving this efficiency.

\subsection{Stability and Generalization}
The consistency of improvement across users (e.g., ranging from +1.99\% to +4.75\% in RecGym Dataset) indicates stable generalization and resilience against user-induced concept drift. Notably, users with lower baseline accuracy tend to benefit more from personalization, especially in QVAR, where inter-user variability was pronounced. suggesting that few-shot adaptation is particularly useful in cases of high inter-user variance or sensor placement discrepancies.

Furthermore, our model adapts without requiring data buffering or batch-based updates. Each incoming labeled sample can be used immediately to refine the classifier, making the approach well-suited for streaming or intermittently labeled data in real-world wearable deployments.

Our system also operates in a fully online manner without requiring batch updates or sample buffering. Each incoming labeled instance can be used immediately for personalization, supporting adaptive learning in real-time wearable settings.

\section{Conclusion}

We introduced a hybrid framework for human activity recognition (HAR) that unifies strong cross-user generalization with fast, on-device personalization. Leveraging few-shot learning on the energy-efficient GAP9 platform, our system adapts rapidly to new users by updating the final classifier of a pretrained model directly on-device, keeping the backbone frozen for efficiency.
Evaluated across three diverse datasets, our method achieved consistent post-deployment accuracy gains of 3.73\%, 17.38\%, and 3.70\%, respectively. These improvements were delivered with sub-millisecond latency and per-update energy consumption of several $\mu$J, demonstrating that practical, real-time personalization is achievable even on constrained embedded hardware.
Our findings reinforce the core insight behind this work: generalization alone is not sufficient for robust HAR in the wild, but when paired with rapid on-device adaptation, it enables scalable, user-aware systems. This work lays the foundation for future HAR applications that must operate continuously across diverse users and changing environments, all while remaining within tight hardware budgets.

\bibliographystyle{ACM-Reference-Format}
\bibliography{papers}

\end{document}